# A STUDY OF GENDER CLASSIFICATION TECHNIQUES BASED ON IRIS IMAGES: A DEEP SURVEY AND ANALYSIS

Basna Mohammed Salih Hasan [a,*], and Ramadhan J. Mstafa [b,c]

[a] Technical College of Informatics Akre, Duhok Polytechnic University, Duhok, Kurdistan Region, Iraq- (basna.mhmed@dpu.edu.krd)

[b] Dept. of Computer Science, Faculty of Science, University of Zakho, Duhok 42002, Iraq

[c] Dept. of Computer Science, College of Science, Nawroz University, Duhok 42001, Iraq (ramadhan.mstafa@uoz.edu.krd)

**ABSTRACT:**

Gender classification is attractive in a range of applications, including surveillance and monitoring, corporate profiling, and human-computer interaction. Individuals' identities may be gleaned from information about their gender, which is a kind of soft biometric. Over the years, several methods for determining a person's gender have been devised. Some of the most well-known ones are based on physical characteristics like face, fingerprint, palmprint, DNA, ears, gait, and iris. On the other hand, facial features account for the vast majority of gender classification methods. Also, the iris is a significant biometric trait, because the iris, according to research, remains basically constant during an individual's life. Besides that, the iris is externally visible and is non-invasive to the user, which is important for practical applications. Furthermore, there are already high-quality methods for segmenting and encoding iris images, and the current methods facilitate selecting and extracting attribute vectors from iris textures. This study discusses several approaches to determining gender. The previous works of literature are briefly reviewed. Additionally, there are a variety of methodologies for different steps of gender classification. This study provides researchers with knowledge and analysis of the existing gender classification approaches. Also, it will assist researchers who are interested in this specific area, as well as highlight the gaps and challenges in the field, and finally provide suggestions and future paths for improvement.

## 1. INTRODUCTION

There are different reasons why identifying gender from soft biometrics is a remarkable and possibly valuable topic to study and research. In the case of looking for a match in an authorization database, one potential use appears. Identifying the gender of a sample may be used to organize the search, which can result in reducing total inspection time. A further potential advantage comes in social situations when it probably becomes beneficial to screen access to a certain location depending on gender, but with no requirement for the user to reveal their identity (Rahim et al., 2021). Gender classification is also essential for a variety of reasons, including the collection of demographic data, marketing research, and real-time electronic marketing. In high-security situations, knowing the gender of those who seek to enter but are not recognized as authorized individuals may be useful information to have (Rai & Khanna, 2012).

Biometrics are based on inherence factors and thus are able to naturally distinguish one individual from another. Biometric features contribute many benefits. For example, they are unique for each person. Also, they are difficult to forget, steal, borrow, share, or observe. In addition, they change constantly and are always accessible. Moreover, they cannot be easily transferred to someone else. Biometrics is often defined as a field of study that focuses on quantifying and studying an individual's unique traits in order to identify or verify that individual's identification. It may be classified into conventional, primary, and soft biometrics: Biometrics as they are now practiced concern physical, behavioral, and biological traits such as facial features, eye, signature, stride, voice, DNA, and fingerprints (Abdelwhab & Viriri, 2018) as shown in Figure 1.

Soft biometrics is concerned with auxiliary features such as gender, ethnic origin, skin color, scars, and height that give some information but are insufficient to properly identify a person. To be recognized as a biometric trait, behavioral or physiological attributes of humans must satisfy the following requirements: 1- Universal: everyone individual has the characteristic. 2- Acceptable: readily accessible when assistance is required. 3-Resistance to circumvention: cheating is difficult. 4- Distinctive: may be used to distinguish between individuals. 5- Permanence: they remain constant throughout time. 6- Collectible: the trait is readily collectible and measurable (Reshma et al., 2018).

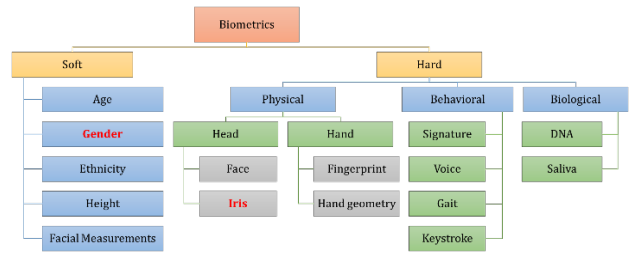

Figure 1. Biometrics types (Abdelwhab & Viriri, 2018).

Recently, iris images have been used extensively to determine gender classification based on iris images. In the human body, only the iris is visible from the outside. As a result of ecological impacts, it cannot readily be changed since it is a specified

---

internal physical characteristic. There is a probability of finding two identical eyeballs of 1 in 10^52. On the other hand, it contains more than 250 points of identification. During fetal development, minute iris patterns and textures are formed, which are phenotypic characteristics, and are not inherited from the gene, which is why identical twins do not have identical eyes. The texture patterns of the eye are not inherited genetically (epigenetic), despite the fact that the color and appearance of the eye are genotypes.

During the sixth week of pregnancy, the iris is formed. A ciliary fold develops by the third month in the iris, and a vascular layer develops by the fourth month in the iris. A sphincter muscle, pupillary membrane, and iris stroma are formed during the fifth month of pregnancy. In the pupillary area, the collarette is very close to the pupil by the end of the eighth month, and the stroma is very thin. It takes one year after birth for the pigmentation process to be complete, even though the patterns and structures are complete during this period (Bartfai et al., 1985). Approximately 18 mm are measured from the front to the back of a newborn's eyeball. Until the age of two, the eye develops to a length of approximately 24-25 mm and then stops growing. The circumference is approximately 38 mm, the diameter is roughly 11-12 mm, the radius is approximately 5.5 mm, the thickness at the pupillary edge is 0.6 mm, and the ciliary boundary is 0.5 mm. One-third of the iris' surface is taken up by the pupil, which measures almost 5 mm in diameter and 2.5 mm in radius. There is a 1.5 mm distance between the collarette and the pupil's edge (Willoughby et al., 2010).

Located between the white sclera and the black pupil of the eye, the iris is a thin circular ring. Iris contains unique and rich details regarding texture, such as patterns, rifts, stripes, filaments, coronals, furrows, minutiae, colors, rings, spots, and recesses. An iris consists of loosely woven tissues. The iris controls the amount of light that reaches the retina by regulating the diameter and size of the pupil, which is a hole in the iris's center (Boyd & Turbert, 2021). Figure 2 shows the anatomy of the eye. The iris, according to the study, stays unchanged over an individual's life. Additionally, iris-based biometrics systems may not be intrusive to their users, which is critical for real-world applications, since the iris is directly observable. A practical and intriguing alternative to permanent personal identity, gender classification has all of these desirable properties (uniqueness, long-term stability, non-invasiveness).

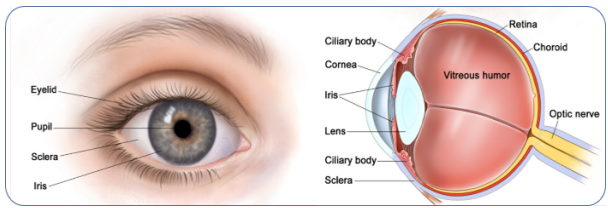

Figure 2. Anatomy of the eye, showing the exterior and interior of the eye, and iris (Willoughby et al., 2010).

There are several approaches to establishing gender identity. Gender recognition utilizing face characteristics employing PCA and LBP, COSFIRE filters, Gabor filters, deep neural networks, and Eigen features. Mobile biometric data, skin texture traits, the whole-body picture, and voice recognition are used to determine the second gender. Men and women have different hairlines, with the former having a higher point and the latter a lower one. Males tend to have thicker ones that rest beneath the orbital rim, while females have higher ones that are arched. The distance between the brows is narrower in men and wider in women (Rekha et al., 2019) giving the former the impression of larger eyes.

Basic procedures for iris-based gender classification are as follows. The first stage involves a process of acquisition; it employs useful sets of data. The following stage is segmentation. From the acquired picture, it pinpoints exactly where the iris is located. The next step is to locate the edge of the pupil. The feature extraction step extracts geometric or texture properties, as well as any other form of features like. The results of this are shown in Figure 3. Training then takes place, during which time these feature parameters are adjusted to their optimal settings. In the prediction phase, we examine the collected information and try to determine the gender of the subject.

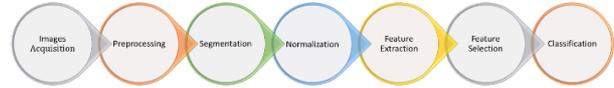

Figure 3. Basic processing stages of iris-based gender prediction (Koklu & Ozkan, 2020).

Accurate gender classification is based on both feature extraction and classification, with feature extraction being essential to successful classification. It calls for the most identifiable characteristics across classes and preserves original characteristics within the same class. Due to its exceptional feature extraction capacity, CNN has recently gained a lot of attention in the machine learning and pattern identification fields. It has achieved the highest levels of performance in image recognition yet and can extract features from images on its own. When originally used to perform these unsupervised tasks, CNN demonstrated considerable gains in picture recognition, and its structure and parameters may be fine-tuned to further raise classification accuracy. In section 4, the fundamentals of convolutional neural networks (CNNs) and their training methodology will be laid out, and various widely used CNN models in the area of computer vision will be presented so that the reader may have a better grasp of CNN-based image categorization.

This article aims to evaluate and compare numerous gender classification techniques, as well as to explain the two methodologies of iris classification, namely the classical approach and the deep learning approach. Also, it provides recommendations and future directions to improve them, as well as to discuss possible areas for future research. Finally, it highlights deep learning's structure adaptable to image structures, and capable of distributing image information and swiftly acquiring it from huge volumes of data. The rest of this paper is divided as follows: Section 2 presents a literature review, section 3 discusses techniques for gender classification, section 4 briefly describes deep learning and explains how CNN works, section 5 provides recommendations, and section 6 provides a conclusion.

## 2. LITERATURES REVIEW

A survey of gender profiling advances utilizing deep learning and machine learning approaches over the last four years is presented in this section. In addition, table 1 displays the results of the papers.

The hybrid CNN-ELM approach for age and gender classification was presented by Duan et al. (2018). An Extreme Learning Machine (ELM) and a Convolutional Neural Network (CNN) were used together to categorize age and gender. ELM classified the intermediate results based on the features extracted from the input images in the hybrid design, which takes advantage of the strengths of CNN and ELM. In order to prevent overfitting, they included a series of cautions, including analyzing the hybrid architecture, creating parameters, and using backpropagation throughout the iteration process. The hybrid structure was confirmed using MORPH-II and the Adience Benchmark. Compared to previous experiments on the same datasets, the hybrid architecture had significantly better accuracy and efficacy results than previous research.

In addition, Manyala et al. (2019) classified near-infrared periocular images according to gender using CNN. Based on near-infrared images of the periocular area, the researchers tested two methods for gender classification based on convolutional neural networks (CNNs). These techniques proceed by identifying and extracting the left and right periocular regions in the first stage. First, periocular images were processed using a pre-trained CNN trained on domain-specific information. In the next step, SVMs were used to predict gender information. Using periocular images, the second strategy was developed using a pre-trained CNN. A public database and two internal databases were used for this analysis. Local binary patterns and histograms of oriented gradient-based methods were used as baselines for determining the efficiency of the suggested techniques. The findings demonstrated that the suggested techniques outperformed the baseline methods in terms of classification accuracy, especially when applied in one of the public datasets that include a substantial number of non-ideal images. Additionally, the suggested techniques consistently outperformed the previous method in terms of accuracy.

As well as, Rattani et al. (2018) ,employed convolutional neural networks to detect gender from ocular images taken with a smartphone. They carried out a detailed investigation on gender prediction using ocular images gathered through front-facing cameras on smartphones. This was a new issue since prior studies in this field had not examined RGB ocular images obtained by cell phones. They achieved the goal via the use of deep learning. For gender classification, pre-trained and bespoke convolutional neural network architectures have been employed. To boost the accuracy rate, multi-classifier fusion was applied. Additionally, a comparative investigation of off-the-self textural descriptors and human ability in gender classification was undertaken.

Moreover, Aryanmehr et al. (2018), with Computer Vision and Biometric Laboratory (CVBL) introduced image processing and biometric research using the IRIS gender identification database. They created an iris image database for gender classification and evaluated it using a new gender classification algorithm. The collection contains iris images of 720 university students, including 370 females and 350 males. Each student had more than six images taken of his or her left and right eyes. Following an examination of the images, three from the left eye and three from the right eye were chosen as the most relevant and added to the database. The proposed database's quality and efficiency were tested using a new approach for extracting Zernike moments from spectral characteristics and two well-known classifiers, SVM and KNN. The results indicated a considerable improvement in gender classification when compared to comparable datasets.

In 2019 Patil et al. (2019), presented gender classification by a multimodal biometrics system from images of the face, iris, and fingerprint. Using the same subject's face, iris, and fingerprints, they may identify the subject's gender. Features were fused to provide robustness in genetic analysis and an accuracy of 99.8% was attained on the multimodal biometric database SDUMLA-HMT, which they suggested to work (Group of Machine Learning and Applications, Shandong University). The results showed that the Multimodal Biometric System's feature level fusion improved gender classification performance significantly.

Additionally, Fang et al. (2019) ,employed a convolutional neural network to classify genetically identical left and right irises; however, the iris identification process focused only on the differences in iris textures, neglecting similarities. They used a VGG16 convolutional neural network to study the link between an individual's left and right irises. Experiments using two different iris datasets demonstrated that when identifying whether two irises (left and right) were from the same or different individuals, a classification accuracy of more than 94 % may be achieved. This work showed that deep learning can detect similarities between genetically identical irises that were previously indistinguishable using classic Daugman's techniques. They believed that this study will pave the way for future research on the association between irises and/or other biometric identifiers of genetically identical or related persons, which may be used in criminal investigations.

In addition, Khalifa et al. (2019) ,classified gender using iris patterns and deep learning. They proposed a technique for robustly identifying the gender of an iris using a deep convolutional neural network. By using the graph-cut segmentation approach, the suggested architecture separated the iris from a backdrop image. The first collection had 3,000 images, 1,500 of males and 1,500 of women. The augmentation strategies used in their study overcame the issue of overfitting and improved the suggested architecture's resistance to simply remembering the training data. Additionally, the augmentation method increased the number of dataset images to 9,000 for the training phase, 3,000 for the testing phase, and 3,000 for the verification phase, while also significantly improving testing accuracy, with the suggested architecture achieving 98.88 %. A comparison of the suggested approach's testing accuracy with that of another previous research utilizing the same dataset was offered.

Furthermore, Rajput and Sable (2020), estimated gender and age using deep learning and the iris of a person. They suggested a deep convolutional neural network for age estimation and pre-trained deep convolutional neural networks AlexNet and GoogLeNet for iris-based gender classification. To extract attributes from iris images, deep learning pre-trained networks were used. Additionally, these attributes were trained and classified using a multi-class SVM model to evaluate the system's performance. The hypothesis that the iris included age and gender information was confirmed by experimental data.

Moreover, Eskandari & Sharifi ( 2019), were tasked with analyzing face-ocular multimodal biometric systems in order to estimate a person's gender. They presented the first study to explore the fusion of facial and ocular biometrics in order to estimate a person's gender using a hybrid multimodal approach. They want to investigate the effect of multimodal biometric systems on gender identification at the score and feature levels. The facial and ocular texture information was extracted through the implementation of a uniform local binary pattern (ULBP) feature extractor. They suggested employing a unique evolutionary technique called backtracking search to choose the most efficient feature sets for both modalities (BSA). On the other side, employing fused facial and ocular attributes and scores, an SVM was employed for classification purposes. The suggested approach was verified utilizing the CASIA-Iris-Distance and MBGC multimodal biometric datasets, taking into account subject-disjoint training and testing. The gender recognition attained revealed the hybrid multimodal face-ocular scheme's superiority over the unimodal face-ocular system used in their investigation for subject gender prediction.

As well as, Tapia and Arellano (2019), investigated the application of a Binary Statistical Features (BSIF) method for gender classification using NIR-captured iris texture images. It followed the same pipeline as iris recognition systems, which involves segmentation, normalization, and classification of the iris. Experiments demonstrated that using BSIF was not easy, since it might result in the creation of artificial textures, which could result in misclassification. To avoid this constraint, a new set of filters was trained using eye images, and alternative sized filters with padding bands were evaluated on a subject-disjoint database. They used the Modified-BSIF (MBSIF) approach. The latter produced more accurate gender classifications (94.6\ percent and 91.33\ percent for the left and right eye respectively). These findings were similar to those obtained using state-of-the-art gender classification techniques.

Additionally, a unique gender-labeled database has been produced and will be made accessible upon request.

Tapia and Perez (J. E. Tapia & Perez, 2019), estimated the gender classification rate using a two-dimensional quadrature quaternionic filter and a subset of the most relevant features from normalized iris images. Instead of using the 1-D log-Gabor encoding method, they encoded the phase information of the normalized images using four bits per pixel using a 2-D Gabor filter and selecting the best bits from the four resulting images (1 real and 3 imaginary). They examined the effectiveness of hand-crafted and automated approaches for selecting and extracting the most relevant features from full iris images, image blocks, and pixel features. Selecting iris blocks and attributes saved computing time and was critical for determining what information, in addition to pixels from the iris, can be collected to determine gender on a fundamental level. The Quaternionic-Code with the complimentary feature selection approach obtained the greatest results on the GFI-UND database, with a left iris detection rate of 93.45 % and a right iris detection rate of 95.45 %, both with 2400 attribute values. They compared their findings to the best previously reported results, as well as to those achieved using convolutional neural network feature extraction.

Then, Tapia et al. (J. Tapia et al., 2019), showed gender classification using images of the periocular iris from mobile phones. They demonstrated the use of Super-Resolution-Convolutional Neural Networks (SRCNNs) to improve the resolution of low-quality periocular iris images cut from subjects' selfie images. They found that when employing a Random Forest classifier, increasing image resolution (2x and 3x) may enhance the sex identification rate. The right eye had the highest identification rate of 90.15 %, while the left eye had the lowest classification rate of 87.15 %. This was obtained by upscaling images from 150x150 to 450x450 pixels. Their findings were comparable to the prior art and showed that when image resolution was raised using the SRCNN, the sex-classification rate improved. Additionally, a new selfie database was constructed using images taken by 150 people using an iPhone X and is accessible upon request.

In 2020 Gornale et al. (2020), employed deep learning to estimate gender using multimodal biometric data analysis. A robust multimodal gender detection approach based on deep features was developed utilizing a commercially available pre-trained deep convolutional neural network architecture based on AlexNet. The suggested model was composed of 20 subsequent layers that included convolutional layers with changing window sizes and fully linked layers for feature extraction and classification. Extensive tests on a similar SDUMLA-HMT (Shandong University Group of Machine Learning and Applications) multimodal database including 15052 images were performed. The suggested approach produced a 99.9 % accuracy rate.

In addition, Cascone et al. (2020), offered pupil size as a soft biometric for identifying age and gender. They conducted comprehensive research with the goal of proving that pupil size and dilation over time may be used to identify individuals by age and gender. To do this, 14 supervised classifiers were used for a gaze analysis dataset. They demonstrated measuring the right and left pupil separately and concurrently, comparing the performance of classifiers, and selecting the poorest and best performers to support prospective fusion procedures. While satisfactory results for age categorization were achieved, with an accuracy range of 79–82 %, the same cannot be stated for gender. This was owing to the fact that this biometric feature is dual in nature. Pupil size may be regarded as a physical characteristic for age categorization and a behavioral characteristic for gender classification. The study's findings indicate a need for more fair and dependable databases.

Moreover, Cantoni et al. (2020), presented demographic identification using pupil analysis. Gender and age are estimated from pupil size as distinguishing features. The data collected during free observation of face images were utilized to train two classifiers (Adaboost and SVM), considering both the best results provided by each classifier and their fusion through weighted means. They discovered that using data on fixations and gaze pathways, they could get more significant findings using pupil size than they could use data on fixations and gaze paths. Pupil Diameter Mean (PDM) has been shown to be the most distinguishing feature across gender and age groups.

In 2021, Khan et al. (2021) ,proposed an authentication technique based on the SVM classification of iris images. It has a great reaction to persistent changes when the Zernike, Legendre, and Gradient-oriented histograms are used. They extracted features from iris images using invariant moments. After extracting the features from these descriptors, the attributes were classified using keycode fusion. SVM was used to classify individuals based on their gender using a fused feature vector. The suggested method was examined using the CVBL data set, and the findings were compared to those obtained using state-of-the-art techniques such as local binary patterns and Gabor filters. The suggested technique achieved a gender classification rate of 98 % with a low computing cost, making it suitable for use as an authentication mechanism.

Moreover, Alghaili et al. (2020) employed deep feature learning to classify individuals based on their covered/camouflaged faces. They suggested a network that combines inceptions with a loss function based on variational feature learning (VFL). The suggested network used the central region of the face to determine the gender of both normal and concealed/ camouflaged faces. The network was taught to focus on the central region of faces, which typically includes the eyes and has a little gap at the top left and bottom right. Five publicly accessible data sets (FEI, SCIEN, AR FACES, LFW, and ADIENCE) were used, and the suggested network outperformed state-of-the-art methods on all of them. In addition, they evaluated their network on a new dataset consisting of masked and covered faces and found similarly positive outcomes.

Additionally, Payasi and Cecil (2021), presented how to classify human gender using radial Support Vector Machines based on LBP and iris features. In their study, they employed a Redial kernel SVM basis classifier for gender identification using Iris crypt densities, Histogram of Oriented Gradients (HOG), and Local Binary Patterns (LBP) as classification features. Along with the Crypt count, their method used features from the Histogram of Oriented Gradients (HOG) to determine the orientation of a human face. They were able to detect the orientation using just HOG features and achieved a gender classification accuracy of 67.85 %. Local Binary Patterns (LBP) were discovered to be different in male and female faces; hence, their experiment included LBP as an additional classification feature; they obtained an 80.55 % classification rate using just LBP features. The densities of iris crypts on the right and left iris were quantified and statistically evaluated. The suggested study revealed that females had a larger iris-crypt density in both of the locations tested, separately and in combination. In circumstances when incomplete eye-iris data is discovered, differences in crypt density may be a significant tool for determining gender. They obtained a gender classification accuracy of 90% using Crypt densities. Their study included all three features and discovered a classification rate of 98.5 % for gender using the Redial kernel Support Vector Machine (SVM) classifier. Their study was conducted using MATLAB 2018b and the FERET standard human face database for recognition.

**2.2. Table** 1. Summary of last works about gender classification using different approaches.

| Ref. | Datasets | Feature Extraction | Classifier | Tasks | Biometric | Contributions | Accuracy |
|---|---|---|---|---|---|---|---|
| (Duan et al., 2018) | MORPH-II, Adience Benchmark | Hybrid CNN-ELM, fuzzy filter | ELM, SVM | Age & Gender | face | Applying the advantages of CNN & ELM | 88.2% |
| (Manyala et al., 2019) | In-house, MBGC portal, IIITD multispectral periocular | CNN, VGG-Face, extract deep feature | SVM | Gender | periocular | Eyebrow traits have a significant role in gender classification. | 94% |
| (Rattani et al., 2018) | FRGC, web images, FERET, images of groups, VISOB subset | CNN, VGG16-19, ResNet, | SVM, MLP, KNN, Adaboost | Gender | ocular | Texture descriptors are evaluated against human performance in gender prediction using deep learning algorithms. | 87% |
| (Patil et al., 2019) | SDUMLA-HMT, dataset FT-2BU sensors | fusion of MB-LBP, BSIF filters | LDA, KNN, SVM | Gender | Face, Iris, Fingerprint | Present a multimodal biometrics-based gender recognition system that integrates characteristics from three distinct biometric modalities, namely the Face, the Iris, and the Fingerprints. | 95.7% |
| (Fang et al., 2019) | CAS-SIAT, CASIA-Iris-Interval | Deep learning, CNN | VGG16, VGG19 | the correlation between the left and right irises | iris | The correlation in iris patterns between people who have a common ancestor or are otherwise closely related | 94% |
| (N. E. M. Khalifa et al., 2019) | GFI (The original dataset consists of 3,000 images) | deep Learning, ReLU, CNNs, GD, back-propagation | CNN | Gender | iris | The use of a deep convolutional neural network for accurate iris gender recognition is shown. | 98.88% |
| (Rajput & Sable, 2020) | collected | Deep CNN for age estimation. And pre-trained deep CNN AlexNet and GoogLeNet for gender prediction | SVM | Gender and Age | | The hypothesis 'the iris has age and gender-related information is proved valid through experimental findings. | 95.34% |
| (Eskandari & Sharifi, 2019) | CASIA-Iris-Distance, MBGC face–ocular | ULBP, BSA | SVM | Gender | face–ocular | Using effective feature subsets of modalities for gender recognition in person-disjoint evaluation. | 89% |
| (J. Tapia & Arellano, 2019) | GFI-UND UNAB-Val | BSIF, MBSIF | Adaboost, LogitBoost, GentleBoost, RobustBoost, LPBoost, TotalBoost, RusBoost. RF, Gini Index, LIB-SVM | Gender | iris | MBSIF and MBSIF histogram are two methods that have been put into practice. | 94.6% for the left eye. And 91.33% for the right eye. |
| (J. E. Tapia & Perez, 2019) | GFI-UND | VGG19, genetic algorithm, Condition Mutual Information Maximization, Random Forest | AdaboostM1, LogitBoost, GentleBoost, RobustBoost, LPBoost, TotalBoost, RUSBoost, Random Forest, SVM | Gender | iris | They looked at the depth and detail of the iris normalized-encoded images and investigated the possibility of using 2D Gabor filters to extract 2D spatial information at a rate of four pixels per bit, as opposed to the more common 1D Gabor filter approach of two pixels per bit. | 93.45% for the left iris and 95.45% for the right iris |
| (J. Tapia et al., 2019) | The innovative INACAP database was obtained for their study. | Super-Resolution Convolutional Neural Networks (SRCNNs) | RF, SVM | Gender | periocular iris | It has been shown that the success rate of gender categorization increases as image resolution is improved by SRCNN. A fresh selfie database has also been developed. | 90.15% |
| (Sharanappa Gornale et al., 2020) | SDUMLA-HMT multimodal biometric database. | Deep convolution neural network architecture based on AlexNet. | Naive Bayes, KNN, SVM, and Decision Tree | Gender | multi biometrics | Using a commercially available pre-trained deep convolution neural network architecture based on AlexNet, we devised a powerful multimodal method to gender detection based on the deep features computed from the input data. | 99.9% |
| (Cascone et al., 2020) | GANT | measurement of its diameter | SGD, SVM, BC, RF, KNN, Adaboost, GB. | Age and gender | pupil | They presented an extended investigation to prove that pupil size and dilation overtime may be | ranging in 79%–82% of accuracy |
| | | | | | | possibly utilized to categorize persons by age and by gender. | |
| (A. R. Khan et al., 2021) | CVBL dataset | Deep belief &(MLP), CNN, CNN& MLP, WSVM | SVM | gender | iris | Gender classification techniques have high accuracy rates as well as low computational complexity | 97.5% |
| (Alghaili et al., 2020) | FEI, SCIEN, AR FACES, LFW, and ADIENCE | NN4 Network with variational feature learning (VFL) loss function | Softmax classifier | Gender | middle part of the faces | Extraction of discriminating visualizations of gender classification with covered or masked faces | 97.22% |

## 3. GENDER CLASSIFICATION TECHNIQUES

### 3.1. Human and computer vision

Recognizing objects is one of the most difficult challenges in computer vision. Computer vision makes use of machine learning methods and algorithms to acquis, image process, recognize, describe, and identify objects based on their size or color, as well as to detect and understand patterns in visual data such as images and videos. By detecting things within its area of view, computer vision mimics human vision. Thus, researchers study the iris from a computer vision viewpoint, but they must always verify how their findings may be affected by other factors. Thus, the physical characteristics of the eye and medical information about it provide the researcher with a broad understanding of the iris as shown in figure 4 (Bhanu & Kumar, n.d.). The iris is composed of two main regions (Angée et al., 2021):

- The pupillary zone is the inner area whose border creates the pupil's boundary.
- The ciliary zone encompasses the remainder of the iris, extending all the way to its start at the ciliary body.
- The collarette is the area of the iris that separates the pupillary and ciliary halves. Typically, it is described as the area where the sphincter and dilator muscles meet.

The iris's surface is highly intricate. John Daugman described the iris's 250 features. The below are the most essential for identification:

- Crypts: atrophy in front and stroma generating the iris's characteristic drawing; they are thin areas in the iris.
- Radial furrows: a series of very fine razor-like nibs that extends from the pupil to the collar.
- Pigment spots: irregular groups of pigment on the iris's surface.

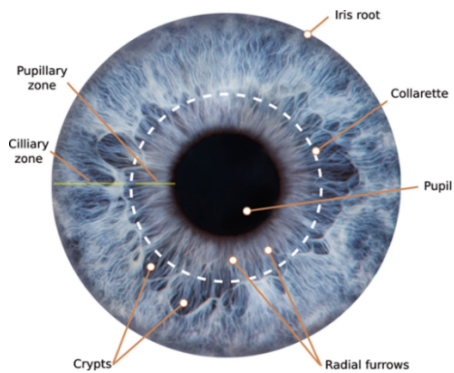

Figure 4. Structure of the iris (Angée et al., 2021).

**3.1.1. Iris Images Acquisition:** Researchers may do a study using their own private image datasets or benchmark databases (produced and made publicly available online by various organizations). Images may be acquired using a camera or other input sensors in the image acquisition module to construct image databases. Moreover, many public iris datasets with male and female pictures are used for gender classification studies. Not only iris datasets but also a facial database with male and female pictures used in gender classification (Hassan et al., 2021).

Another new benchmark data set is Adience, which is being used to classify genders. This data set is considered to be wild and large-scale for face soft-biometric classification. The Adience data collection as its whole comprises 26,000 images of 2284 individuals. The benchmark is comprised of images published to Flickr from smartphone devices. This method of capturing and uploading images without filtering them accurately replicates real-world face images. As a result, the Adience data set contains images with considerable variance in head posture, illumination quality, and noise. Gender information extraction is subject to variation owing to variations in lighting, stance, expression, age, and ethnicity. Similarly, throughout the image capture process, variables affecting image quality such as dithering, noise, and poor resolution contribute to the difficulty of image analysis (Duan et al., 2018).

**3.1.2. Images Segmentation and Normalization:** This module's major objective is to enhance images to extract relevant information from them. There are other pre-processing methods available, such as histogram equalization for contrast enhancement (Kumari et al., 2012). This approach determines the most frequent intensity value in the image histogram and changes the image's global contrast accordingly. The multiscale retinex (MSR) method (Juefei-Xu & Savvides, 2012) is a variant of the single-scale retinex algorithm in that it employs the combined output of multiple smoothing kernels of varying sizes as center-surround image filters to handle varying lighting effects.

The final image has a high perspective, while the algorithm specifies and reacts to the iris borders in lower dimensions. As a result, the imaging tool manually crops the eye image and separates it from the entire image in an area nearly equal to the size of the two eyelids indicated in Figure 6 and the distance between the iris and the pupil and the iris and the crust. Finally, Daugman's approach is used to segment the region around the iris and pupil on the cropped image (Batchelor, 1980).

Iris segmentation is the technique of detecting and recognizing the iris and pupil boundaries of an iris in an eye image. This approach enables the extraction of exact and precise iris features for personal identification. To put it another way, segmentation's major objective is to remove ineffective regions, such as those outside the iris (eyelids, eyelashes, and skin). The quality of the ocular image is critical to the segmentation operation's success. The segmentation process determines the iris and pupil boundaries, which are then transformed into an appropriate template during the normalization (Mabuza-Hocquet et al., 2018) step as shown in figure 6.

After the successful segmenting step, the segmented iris is changed to a precise dimensional pattern to facilitate feature extraction. The normalizing approach creates iris sections with equal fixed dimensions, resulting in two images of the same iris taken under different lighting conditions exhibiting disparate features at the same spatial location. Discontinuities in the dimensions of the normalized iris may occur as a consequence of pupil expansion, which leads to iris stretching. Additionally, some factors such as the quantity of light falling on the eye, the imaging distance, the head tilt, and the camera rotation may all contribute to erroneous iris normalization processes (Daugman, 2009).

Iris detection depends on the great contrast between the pupil and the iris, as well as the iris and the sclera. Daugman (2009), the pioneer in iris recognition, presented an elegant and efficient method, to locate the circular boundary parameters - the circular integro-differential operator as specified in Equation:

$$\max_{(r,x_0,y_0)} \left| G_\sigma(r) * \frac{\partial}{\partial r} \oint_{r,x_0,y_0} \frac{l(x,y)}{2\pi r} ds \right| \quad (1)$$

where: $l(x,y)$– image pixel color, $G_\sigma(r)$ the Gaussian smoothing function with scale $(\sigma)$ is given by the following formula:

$$G_\sigma(r) = \frac{1}{\sqrt{2\pi\sigma}} \exp\left(-\frac{(r-r_0)^2}{2\sigma^2}\right) \quad (2)$$

Where $r$ – radius and $\sigma$ – contour given by circle with $r, x_0, y_0$ parameters.

The normalizing process; is done through various approaches. As J. Daugman proposed the cartesian to polar coordinate transform, it transformed the area of interest into a rectangle and introduced an intriguing normalizing approach. Thus, we may more easily extract the frequency information contained inside the limited zone. Daugman's rubber-sheet model of iris normalization can be stated as follows:

$$I(x(r,\theta), y(r,\theta)) \to I(r,\theta) \quad (3)$$

Where

$$\begin{aligned} x(r,\theta) &= (1-r)x_p(\theta) + rx_i(\theta) \\ y(r,\theta) &= (1-r)y_p(\theta) + ry_i(\theta) \end{aligned} \quad (4)$$

Where $x_p$ the radius of pupil, $x_i$ the radius of iris, $x_p(\theta), y_p(\theta)$ the pupillary boundaries, and $x_i(\theta), y_i(\theta)$ the limbic boundaries in the direction y. as shown figure 5.

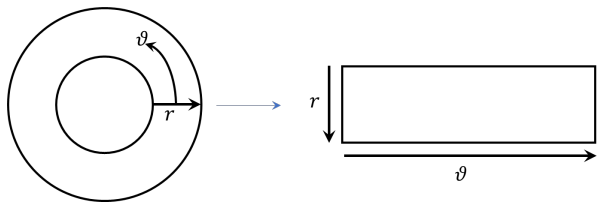

Figure 5. Daugman normalization method.

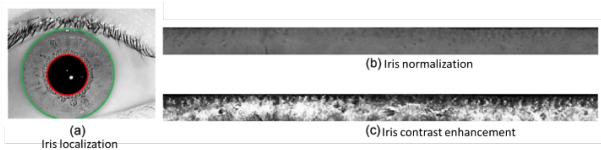

Figure 6. Preprocessing of iris image (Daugman, 2009).

**3.1.3. Feature Extraction:** Throughout the processing of each iris image, several iris attributes are extracted. The radius and center of the iris and pupil are usable for segmentation and are saved as "geometric" attributes. The iris image is recorded as an array of complex integers with rows and columns during the iris encoding phase. Various techniques are used to extract iris features; these techniques differ in their method of feature extraction. Finally, data are classified using classification algorithms. After segmenting the iris area from the eye image, it is transformed into fixed dimensions to facilitate comparisons. They have a wide variety of techniques for extracting features from normalized iris images, including the Gabor filter, Wavelet transforms, and Hilbert transforms (Kazakov, 2011).

**A. Geometric Features**

Geometric attributes are those that relate to the iris's dimensions. These attributes are extracted at the segmentation phase of iris image processing and consequently do not include any information about the real texture. All of these attributes are measured in pixels squared. The geometric features are as follows (Hollingsworth et al., 2009; I. A. Khalifa et al., 2019):

- The center of the iris in x coordinates.
- The center of the iris in y coordinates.
- The center of the pupil in x coordinates.
- The center of the pupil in y coordinates.
- The distance between the iris and the center of the pupil (X-coordinates)
- The distance between the iris and the center of the pupil (Y-coordinates)
- The distance between the center of the iris and the center of the pupil.
- Pupil area.
- Iris area alone, the difference between iris area and pupil area.
- The total area of the iris and pupil.
- Ratio of iris area and pupil area.
- Radius of the iris.
- Radius of the pupil.

**B. Texture Features**

The texture features are those that were extracted from the iris image during the image processing's encoding step. The iris is first represented as an array of complex integers with m rows and n columns during the encoding process. The array's rows correspond to the iris's concentric rings. Each column represents a distinct angle from the iris's center. Thus, crossing a row is analogous to drawing a circle around the iris that is concentric with the pupil and iris limits and lies in between them. Traversing along a column is analogous to keeping a constant angle from the pupil's center and traveling outward from the pupil-iris border to the iris's outer edge. The texture feature is calculated information from each local iris area (Luo, 2012).

**C. Statistical Features**

The following features are included in the iris image. Two types of statistical features are calculated: angular and radial. A movement in the radial direction is characterized by the transition from the pupil of the iris to the border between the iris and the sclera. Means, medians, and standard deviations are calculated based on the statistical features (ACAR, 2016).

- Mean for each row.
- Median for each row.
- Standard deviation for each row.
- Mean for each column.
- Median for each column.
- Standard deviation for each column.

**3.1.4. Feature Extraction Technique:** Several feature extraction techniques from an iris image have been employed. The feature method of extraction is important, it converts a two-dimensional image to a collection of mathematical parameters. The texture of the iris has significant unique features. Different algorithms are used to extract these features. As a result, it is important to investigate representation techniques capturing the iris's underlying local information (Hassan et al., 2021; Rekha et al., 2019).

The feature extraction approach is used to minimize the number of features required to explain the massive amount of information contained in an iris pattern. The reduction in recognition time and rate for two iris templates is due to the use of accurate feature extraction algorithms. Historically, researchers have used a variety of ways to increase the accuracy level. To produce a more accurate iris identification system, a combination (fusion) of homogeneous or heterogeneous iris image features such as Gabor filters (1-D Gabor filter, 2-D Gabor filter), Local Binary Pattern (Local Ternary Pattern) and their variations have been used. Despite improvements so far, the issue of determining the relevance of the features utilized for classification continues to be a challenge (M. T. Khan et al., 2013; J. Tapia & Arellano, 2019; J. E. Tapia & Perez, 2019).

### A. Gabor Filter

Gabor filters have historically been used to segment and classify textures, as well as to recognize fingerprints and faces. Other Gabor filter variants for eye identification include Ring Gabor Filters (RGF) and Circular Gabor Filters (CGF) (CGF). Gabor filters are characterized as having frequency and orientation selectivity. Gabor filters are made up of two components: a lowpass filter and a band pass filter which is defined as follows (Vijayalaxmi & Rao, 2012):

Real part

$$g_{\lambda,\theta,\sigma,\gamma}(x,y) = \exp\left(-\frac{x'^2+\gamma^2 y'^2}{2\sigma^2}\right)\cos\left(2\pi\frac{x'}{\lambda}+\varphi\right) \quad (5)$$

Imaginary part

$$g_{\lambda,\theta,\sigma,\gamma}(x,y) = \exp\left(-\frac{x'^2+\gamma^2 y'^2}{2\sigma^2}\right)\sin\left(2\pi\frac{x'}{\lambda}+\varphi\right) \quad (6)$$

$$x' = x\cos\theta + y\sin\theta \quad (7)$$

$$y' = -x\sin\theta + y\cos\theta \quad (8)$$

The $\theta$ indicates the Gabor filter's selective orientation with regard to the vertical axis. $f_0$ is the Gabor filter's selective frequency along the $x_0$ axis. The $\sigma$ is the Gaussian function's standard deviations along the x' and y' axes, respectively. (x, y) the spatial domain's center of the respective field.

There are many texture analysis filters, but a Gabor filter is one of the most popular. It determines if an image has certain frequency content in certain directions centered around a particular point. Extracting useful information from a picture may benefit from using Gabor filters with varying frequency and orientation. Two-dimensional Gabor filters have several applications in image processing, including feature extraction in areas including texture analysis, segmentation, document analysis, edge detection, image coding, and image representation. This method improves the aesthetic attractiveness of the assembled texture pictures by optimizing resolution in both the time and space domains (M. T. Khan et al., 2013).

### B. Discrete Wavelet Transform (DWT)

DWTs use wavelet scales and translations that can be tailored in response to different instructions (Rinky et al., 2012). Signal processing and data compression can be decomposed on the basis of wavelets. The DWT method uses wavelet decomposition to construct wavelets that are orthogonal to one another. The nonzero portions of these operations are not present throughout the entire signal length, which is in contrast to sinusoidal basis functions (Brifcani & Al-Bamerny, 2010).

It is possible to enhance the set of features extracted from an iris image by applying two-dimensional DWT (Discrete Wavelet Transform) based features. It can extract features from a set of data with both spatial and frequency resolutions provided by the wavelet transform. LL, HL, LH, and HH are approximate, horizontal, vertical, and diagonal subsampled images. In high-pass and low-pass filters, there are four kinds: HH stands for a high-pass image in both directions; (ii) LH stands for a low-pass image in the vertical direction, but a high-pass image in the horizontal direction; (3) HL indicates a high-pass image in the vertical direction, but a low-pass image in the horizontal direction; and (4) LL indicates a low-pass image in both directions.

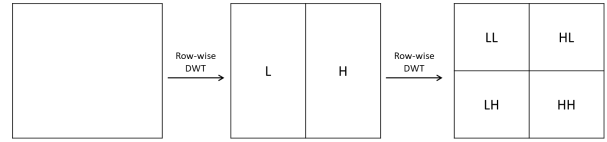

Figure 7. First level of decomposition (Bansal et al., 2012).

The original image of size (NxN) divides into four separate images, each with a distinct frequency component, that make up the original image, which is divided into four smaller images of equal size (N/2 x N/2). Figure 7 illustrates one decomposition process in block diagram form. An image may be decomposed several times using DWT (Bansal et al., 2012).

### C. Discrete Cosine Transform (DCT)

An image in the frequency domain has several characteristics compared to a signal in the spatial domain when it is converted using DCT. A good de-correlation and energy concentration could be demonstrated in the frequency domain of the image. Image data is de-correlated using the DCT. Discrete cosine transformations allow the effective encoding of transform coefficients. Using DCT, you can enhance iris recognition performance by allowing it to be implemented more quickly (Feng & Wu, 2008).

### D. Haar Wavelet Transform (HWT)

An interesting way to extract features from transformation matrices is to use rows as representative samples of the ever-finer resolution, which is a very good and instructive technique. To decompose images, HWT employs low-pass and high-pass filters, respectively, first on the colors of the images and subsequently on their rows. Using the sum and difference of neighboring items, HWT calculates features. It operates on nearby elements first horizontally, then vertically. First applies in the image column then in image rows one by one. The four sub-bands of HWT are designated as follows: LL1, HL1, LH1, and HH1. To generate a fine-grained image, up to four stages of decomposition must be achieved (Podder et al., 2022). The Haar wavelet is constructed from the multiresolution analysis (MRA), which is generated by the Haar wavelet function:

$$\psi(t) = \mathbf{1}_{[0,1/2]}(t) - \mathbf{1}_{[1/2,1]}(t) \quad (9)$$

$$= \phi(2t) - \phi(2t+1) \quad (10)$$

### E. Local Binary Pattern (LBP)

Grayscale changes are responded to by the LBP local texture operator. Based on local community texture descriptions, LBP is derived from local texture descriptions. Local texture patterns are output as binary codes by the LBP operator threshold. In domains requiring rapid feature extraction and texture classification, LBP excels due to its discriminative strength and computational simplicity. Some of the domains where this technique has become more popular include iris identification, visual inspection, image retrieval, remote sensing, face recognition, environmental modeling, and outdoor scene analysis (Ramón-Balmaseda et al., 2012). The equation to calculate the generalized LBP:

$$LBP_{P,R} = \sum_{p=0}^{p-1} s(g_p - g_c) 2^p \quad (11)$$

where, $g_c$ is the gray level of $p$ neighbours $g_p$ (p = 1, 2, ⋯, p − 1).The function $s(x)$ is defined as:

$$s(x) = \begin{cases} 1, x \geq 0 \\ 0, x < 0 \end{cases} \quad (12)$$

**F. Principal Component Analysis (PCA)**

An important aspect of PCA is its ability to compress high-dimensional data to a lower-dimensional format that can be analyzed, visualized, and featured extracted in data analysis. Principal components are calculated by taking a large number of correlated variables and converting them into a smaller number of uncorrelated variables. By estimating the degree of coupling between an image from a training set and an approved image using this principal component, all the information contained in a dataset can be reconstructed. As a result of PCA, non-correlations between variables are reduced to a smaller number of uncorrelated variables. Variables that are not correlated are called "principal components". A basic principle of the data is revealed by these components, which constitute the biggest variance possible ( Kumar et al., 2011).

### 3.2. Machine Learning

After successfully features extraction step, the features are selected to complete classification.

**3.2.1. Feature Selection:** Feature selection is one of the techniques in machine learning in which a subset of the available features from the data is chosen for use by a learning algorithm, such as classification. The most often used strategy is feature selection based on information acquired. This method selects a feature based on the largest amount of information that can be gained from a partition based on the feature (Mabuza-Hocquet et al., 2018).
The length of the feature vector may significantly increase the computational complexity and expense. Sparse representation may aid in identifying discriminating features in big feature sets (S. Kumar et al., 2017). Even with this optimal number of features, it would be able to achieve the same degree of accuracy as with complete features. Thus, feature selection is the process of identifying the optimal features that minimize computation while maintaining high accuracy. Reduced processing will surely speed up the identification process (Adekunle et al., 2020).

**3.2.2. Classification:** As part of machine learning, classification is one of the most fundamental tasks, which consists of finding out which of a set of labels a new sample belongs to, a problem such as selecting from a dataset of images with a variety of categories. The goal is to train a model that can correctly classify unseen images based on the observations in the dataset. The categorization process employs a number of classifier algorithms. Various categorization methods are discussed in each article on gender recognition. There are several categorization methods used, including C4.5 Decision Trees, KNNs, Adaboosts, Random Forests, Gini Indexes, SVMs, Multi-Layer Perceptions, and Optimized Incremental Reduced Error (J. E. Tapia & Perez, 2019)(Cascone et al., 2020) (Rattani et al., 2018; Sharanappa Gornale et al., 2020; J. Tapia et al., 2019; J. Tapia & Arellano, 2019).
The objective of labeled dataset training is to obtain high performance on the test dataset after training a neural network on the training dataset, and the training dataset is typically separated into training and test datasets (and may also include a validation dataset). By comparing new images to the pre-trained model, you can categorize them within the same data distribution space. Neural networks can be applied to a wide variety of applications that classical machine learning methods cannot effectively or efficiently address. This is one of the most exciting aspects of deep learning. The next section will cover the architecture of CNNs and, its models in this area (N. E. M. Khalifa et al., 2019).

### 3.3. Deep Learning by Convolutional Neural Networks

Deep learning-based object recognition does not need any predefined features. The most often used deep learning algorithms are based on convolutional neural networks. Convolutional neural networks are a kind of deep neural networks, which are artificial neural networks having several layers connecting the input and output. A neural network is a kind of computer system that is inspired by the biological neural network found in the brain. One of the primary targets of deep learning algorithms is to detect in real time many layers of distributed impersonations. Deep learning is an area of machine learning algorithms. To derive abstractions at a high level from data. CNNs employ very less pre-processing compared to other image classification techniques. This implies that the network learns to optimize the filters (or kernels) by automatic learning, while in conventional methods these filters are hand-engineered (Fang et al., 2019).

**3.3.1. CNN Structure:** CNNs are multilayer perceptron that are optimized for identifying 2D forms and can be used to map the original input to the desired output. Each neuron in a CNN is linked to the neuron in the preceding layer's local area, lowering the number of weights in the network. CNNs employ a hierarchical connection structure. In other terms, a CNN is made up of layers of components. As seen in Figure 8, they consist of convolutional, pooling, and fully linked layers, as well as an output layer. Convolutional and pooling layers alternate as the first few layers of a conventional CNN, followed by the fully connected layer. Classification results are generated by the final output layer (By, 2018; Kim, 2020).

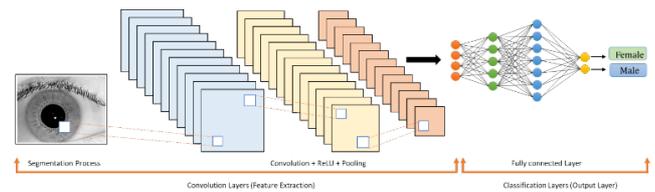

Figure 8. Structure of cnn (Sharma et al., 2018).

**A. Convolutional Layer**

CNNs are built on convolutional layers. With a convolutional kernel of a certain size, a tiny area of the image is convolved. Weight matrices are learned by convolutional kernels. Feature maps are convolved using the output of the convolutional layer and an activation function. A feature map may be used as input for the next convolutional layer. After numerous convolutional layers are layered one after another, more advanced features can be recovered. Consequently, layers of convolutional layers can be combined to obtain more nuanced features. Since the number of parameters in the network does not increase dramatically as the number of convolutional layers increases, the number of neurons in each feature map shares a convolutional kernel's weight. Therefore, this concept may ease the process of building a more complex network (Li et al., 2021).

**B. Pooling Layer**

Pooling layers are usually added after convolutional layers. Maximum, average, and random pooling are the most common pooling layers. Assigning values to neurons based on probability distributions is known as random pooling. Maximum pooling determines the maximum value of neighboring neurons, whereas average pooling determines the average value of neighboring neurons. Overlapping pooling and spatial pyramid pooling are two types of pooling that are often better than standard pooling. In any pooling layer type, features are collected but not oriented toward their placement, allowing the network to continue to learn useful features regardless of changes in input data quantity. It also reduces the spatial dimensions of feature maps while maintaining critical information, thus significantly reducing network training computation, in contrast to a pooling layer which increases the number of feature maps in the preceding layer (Agbo-Ajala & Viriri, 2021).

### C. ReLU Layer

ReLUs employ non-saturating activation functions as rectified linear units. An activation map is essentially empty when negative values are set to zero. In addition, it affects the convolution layers' receptive fields without adding any nonlinearity to the decision function. Other functions, such as the saturating hyperbolic tangent, can also be employed to improve nonlinearity. Since ReLU provides many times faster training than other functions without sacrificing generalization accuracy significantly, it is often preferred over other methods. (Krizhevsky et al., 2012).

### D. Fully Connected Layer and Output Layer

Several hidden layers make up a completely linked layer. A hidden layer is composed of many neurons connected to those in the previous layer. Convolutional and pooling layers produce one-dimensional (1D) feature vectors as inputs for a fully connected layer. In a fully connected layer, these features are translated to a linearly separable space so that the output layer can coordinate with them. By using a classification function, classification results are typically output at the output layer. CNNs are currently equipped with Softmax and SVM functions (Sharma et al., 2018).

### E. Loss Functions

Activation and loss functions play an important role in CNNs. It is common for activation functions to be nonlinear, which allows the network to learn nonlinear mappings layer by layer. There are three common activation functions: sigmoid, Rectified Linear Unit ReLU (Hara et al., 2015), and Maxout (Hinton et al., 2012). Discrepancies between a model's anticipated and actual values are quantified using a loss function, also known as a cost function or an objective function. Overfitting may also be reduced by adding additional regularization units to the loss function, such as L1 and L2. As constraints on a few loss function parameters, L1 and L2 can be viewed as regularizations AlexNet model.

Computer vision researchers consider AlexNet to be one of the most significant studies in the field, as it inspired the use of CNNs and GPUs to accelerate deep learning. A total of eight layers are present in this architecture, including five convolutional layers and three fully coupled layers. A ReLU is used instead of a tan function in AlexNet. In AlexNet, neurons are distributed across two GPUs, enabling multi-GPU training. Not only does this allow for the training of a larger model, but it also reduces training time. Overfitting is less likely to occur in models with overlapping pooling, as well (Szegedy et al., 2014).

- *VGGNet Model*

An example of a sophisticated CNN architecture is VGG, which stands for Visual Geometry Group. In VGG-16 or VGG-19, the "deep" corresponds to the number of convolutional layers, which are 16 and 19 respectively. A ground-breaking object recognition model is based on the VGG architecture. In addition to excelling on many tasks and datasets beyond ImageNet, the VGGNet was developed as a deep neural network. Architecture is still popular today as one of the most popular methods for recognizing images (Gu & Ding, 2019).

- *GoogLeNet Model*

The problem of overfitting may arise if a network has many deep layers. GoogleNet architecture consists of multiple-sized filters that can operate on the same level to solve the overfitting problem, according to the authors of the research paper going deeper with convolutions. Instead of deepening the network, this idea widens it. There are 22 layers in the GoogleNet Architecture, including 27 pooling layers. It consists of nine linearly stacked inception modules. A global average pooling layer (Wu et al., 2017) connects the ends of the inception modules.

## 4. SUGGESTIONS AND RECOMMENDATIONS

Our literature review focuses on the several steps involved in iris-based gender classification. Each step includes a thorough study of the various approaches. Each step has a number of factors that impact classification accuracy; as below:

➢ During the image capture process, factors affecting image quality such as dithering, noise, and poor resolution contribute to the difficulty of image analysis. As a fundamental advance in computer vision, ultra-resolution has been an attractive but challenging solution to this issue in both general imaging systems and biometric systems. However, there is a fundamental distinction between the motives for conventional ultra and those necessary for biometrics. The former tries to increase the scene's visual clarity, while the latter, more importantly, aims to improve classifier identification accuracy by employing special features of observed biometric traits.

➢ A digital image can be obtained by measuring the movement of the iris and pupil of the human eye without, however, affecting the quality of the image due to light-induced reflexes. Czajka and Becker, (2019), suggested employing NIR illumination to acquire images of one eye's pupil while lighting the other eye with a visible-light pulse. This method extracts what are known as "dynamic features (DFs)" from the iris. This new approach suggests extracting information about the human eye's response to light and using it for biometric identification purposes. The findings indicate that these traits are discriminating. Because these DFs can only be derived from live irises, the suggested technology has the potential to be "fraud-proof."

➢ In addition, Preprocessing techniques are used to reduce all sorts of noise from the acquired iris images. Because of the iris's challenges, there has been an increase in false-positive and false-negative mistakes as a consequence of dark region, pupil expansion, the pixels being alike to the iris pixels, sagging of the eyelids due to aging, lens wear, and reflection from eyeglasses. Due to the fact that they hide textural aspects of the iris area, these problems need more research. Thus, when the number of classes becomes exceptionally big, further research will be necessary to determine the breadth and methods of network processing.

➢ Even though iris segmentation is not a critical step in the iris production process, it is nonetheless interesting to learn more about. Occlusion by the eyelid and eyelashes, blurriness, poor lighting, reflections, off-angle pictures, and a lower intensity differential between the pupil and iris are all variables that contribute to the challenge of separating the iris from an eye image. Numerous methods may be enhanced to better handle unrestricted circumstances.

➢ Iris segmentation or localization is an important subject of investigation in iris biometrics, since incorrect localization cannot be remedied at a later stage, which may result in misleading classification. The iris is normally located by detecting two circular boundaries: the limbic contour (iris outside border) and the pupil, as well as two parabolas representing the lower and upper eyelids. Iris localization methods based on the assumption that both boundaries are circular such as Daugman (JG Daugman - US Patent 5 & 1994, n.d.) and Wildes (Wildes, 1997)) often fail when dealing with noisy off-angle images.

➢ In restricted or ideal settings, the image is captured in a controlled environment with an attentive subject. In unconstrained or less-than-ideal settings, however, the image is captured when the subject is reasonably relaxed. As a result, the image quality is poor in the latter situation owing to produced noise such as fluctuations in reflection or light, occlusion of the eyelids and eyelashes, defocus blur, motion blur, low resolution, and a periocular area composed of skin and other features such as birthmarks.

➢ In the ciliary and pupillary zones of the iris contains a wealth of visible traits. The effectiveness of feature extraction is contingent upon accurately representing these textural properties that are necessary for substantially distinguishing the iris throughout the classification phase. Vector features may be classified generically into phase, zero-crossing, texture, and seed points. A few studies used Wavelet transform-based features, Curvelet transform-based features, Energy sum-based features, SIFT-based features, SURF-based features, and Deep features. In addition, the size of the feature vector may significantly increase the computational complexity and expense. Sparse representation may aid in identifying discriminating characteristics in big feature sets. Even with this optimal number of features, it would be able to achieve the same degree of accuracy as with complete features. Thus, feature selection is the process of finding the optimal features that minimize computation while maintaining high accuracy. Reduced processing will definitely accelerate the recognizing process.

➢ Many feature selection approaches have been used in pattern recognition, image processing, and computer vision, including Principal Component Analysis (PCA), Linear Discriminant Analysis (LDA), and Independent Component Analysis (ICA). These strategies did not provide the most useful or optimal features. As a result, metaheuristic optimization approaches for feature selection must be included. The complexity and computing expense of the classifier may be lowered by feature selection by condensing the number of characteristics to be employed into quantifiable forms while retaining an acceptable level of recognition accuracy. Obtaining an optimum feature subset in feature selection is often intractable, and a large number of additional feature selection issues are demonstrated to be the non-deterministic polynomial hard problem (NP). A robust optimization methodology, such as metaheuristic optimization, is needed for this kind of problem.

➢ A number of metaheuristic optimization algorithms, inspired largely by nature, can be used in future research to select the most relevant and discriminant features prior to fusion, including firefly optimization, cuckoo optimization, artificial bee colonies, and ant colonies, gravitational search optimization, and bacterial optimization. The iris recognition system will be able to lower misclassification rates by reducing unnecessary features during the classification step.

➢ As soft biometrics are sensitive to changes in lighting, expression, and position, deep learning can be used to preprocess and extract features. Furthermore, new soft biometric features can be implemented, including face distance measurement and head-to-body ratio. CNNs, a form of a deep learning network that performs direct convolutions on pixels in order to deliver abstract information, offer many advantages over shallow structural models. This feature extraction technique is more generalizable and can be applied to various situations. A CNN is capable of quickly acquiring image information from enormous amounts of data and distributing it across a large area. Due to CNNs' structure, they can efficiently solve complex nonlinear problems (for example, rotating and translating an image). As a result of sparse connections, weight sharing, and spatial subsampling, CNNs provide a network structure that is more flexible to image structure changes.

➢ The segmentation step separates the iris texture from the surrounding texture. Deep learning approaches to aid in the generation of precise and rapid segmentation. The texture descriptors for the iris area are then represented using the appropriate approaches for feature extraction based on deep learning. Multiple attributes give sufficient information about the iris image's characteristics. Additionally, CNN approaches rely heavily on massive and expensive computations. As a result, the most advantageous approaches in some ways are effective owing to the careful selection of the most advantageous aspects. Swarm intelligence techniques and genetic algorithms may be used to optimize the iris region's attributes. Classification approaches may aid the system in making the appropriate judgment to achieve classification (Banerjee & Mery, 2015). In general, the SVM classifier outperforms the artificial neural network classifier. However, the performance efficiency of artificial neural network systems may be improved further by optimizing parameter settings. Additionally, various hybrid approaches such as classifiers based on fusion methods may be used to improve the performance of classification techniques.

➢ The solution lies in developing a new classifier that utilizes discriminative characteristics while maintaining the same capabilities as full-connection layers or SoftMax classifiers (Alghaili et al., 2020). According to A. R. Khan (A. R. Khan et al., 2021), a hybrid classifier combining CNN and Support Vector Machine (SVM) yielded much better results than just a CNN. Since SVMs are so complex, it is critically important to identify alternative classifiers that require fewer tuning parameters, perform well in classification, and are capable of handling the same tasks. There are three prominent classification algorithms at the moment: SVM, Naive Bayes (Sharanappa Gornale et al., 2020), and Extreme Learning Machine (ELM), with ELM proving to be the most efficient and quick approach due to its rapid training speed, high generalization performance, and minimal human involvement (Duan et al., 2018). To handle pattern recognition tasks efficiently, ELMs, enhanced ELMs, and their combinations have been extensively used.

➢ Typically, the eye's problems are reflected in its iris shape, pattern, and texture because they influence the information code included in its structure and construction. Thus, to determine which factors, influence the success rate of iris classification and identification, researchers need to examine the iris's natural characteristics. They should be aware of these, as well as other environmental and geometrical changes (with the light effects as their main factor). Any change in the pattern information supplied by the iris to the automated machinery is significant and warrants investigation. This is because any imperfection that affects the iris and alters or deforms its shape or pattern will impact the outcomes.

## 5. CONCLUSION

Automated gender classification has grown in importance through time and has developed into a significant area of research. Various researchers have invested great effort and generated high-quality research in this field. Furthermore, this field has enormous potential due to its usefulness in a variety of fields such as monitoring, surveillance, commercial profiling, and human-computer interaction. Security applications are critical in this field. Different commercial applications, including human-computer interaction and computer-aided physiological and psychological analysis. Researchers in the area of iris biometrics have developed a variety of techniques for determining a person's gender via iris biometric information.

The purpose of this study is to summarize the majority of the papers conducted by researchers over the last four years by looking into and comparing the many methodologies, algorithms, and findings made. Our work will benefit someone who is inexperienced with gender classification by iris biometrics and want to get a rapid understanding of iris characteristics, and the various methods available to do this.

## REFERENCES

Abdelwhab, A., & Viriri, S. (2018). A Survey on Soft Biometrics for Human Identification. *Machine Learning and Biometrics*. https://doi.org/10.5772/INTECHOPEN.76021

ACAR, E. (2016). EXTRACTION OF TEXTURE FEATURES FROM LOCAL IRIS AREAS BY GLCM AND IRIS RECOGNITION SYSTEM BASED ON KNN. *European Journal of Technic*, 6(1), 44–52.

Adekunle, A., A, Y., Aiyeniko, A., O, O., Eze, E., O, M., & O.D, A. (2020). Feature Extraction Techniques for Iris Recognition System: A Survey. *International Journal of Innovative Research in Computer Science & Technology*, 8(2), 37–42. https://doi.org/10.21276/ijircst.2020.8.2.5

Agbo-Ajala, O., & Viriri, S. (2021). Deep learning approach for facial age classification: a survey of the state-of-the-art. In *Artificial Intelligence Review* (Vol. 54, Issue 1). Springer Netherlands. https://doi.org/10.1007/s10462-020-09855-0

Alghaili, M., Li, Z., & Ali, H. A. R. (2020). Deep feature learning for gender classification with covered/camouflaged faces. *IET Image Processing*, 14(15), 3957–3964. https://doi.org/10.1049/iet-ipr.2020.0199

Angée, C., Nedelec, B., Erjavec, E., Rozet, J. M., & Taie, L. F. (2021). Congenital Microcoria: Clinical Features and Molecular Genetics. *Genes 2021, Vol. 12, Page 624*, 12(5), 624. https://doi.org/10.3390/GENES12050624

Aryanmehr, S., Karimi, M., & Boroujeni, F. Z. (2018). CVBL IRIS Gender Classification Database Image Processing and Biometric Research, Computer Vision and Biometric Laboratory (CVBL). *2018 3rd IEEE International Conference on Image, Vision and Computing, ICIVC 2018*, 433–438. https://doi.org/10.1109/ICIVC.2018.8492757

Banerjee, S., & Mery, D. (2015). Iris Segmentation Using Geodesic Active Contours and GrabCut. *Lecture Notes in Computer Science (Including Subseries Lecture Notes in Artificial Intelligence and Lecture Notes in Bioinformatics)*, 9555, 48–60. https://doi.org/10.1007/978-3-319-30285-0_5

Bansal, A., Agarwal, R., & Sharma, R. K. (2012). SVM based gender classification using iris images. *Proceedings - 4th International Conference on Computational Intelligence and Communication Networks, CICN 2012*, 425–429. https://doi.org/10.1109/CICN.2012.192

Bartfai, A., Levander, S. E., Nybäck, H., Berggren, B. M., & Schalling, D. (1985). Smooth pursuit eye tracking, neuropsychological test performance, and computed tomography in schizophrenia. *Psychiatry Research*, 15(1), 49–62. https://doi.org/10.1016/0165-1781(85)90039-3

Batchelor, B. G. (1980). Book Review: Digital Image Processing. In *The International Journal of Electrical Engineering & Education* (Vol. 17, Issue 3). https://doi.org/10.1177/002072098001700324

Bhanu, B., & Kumar, A. (n.d.). *Advances in Computer Vision and Pattern Recognition Deep Learning for Biometrics*.

Boyd, K., & Turbert, D. (2021). Eye Anatomy: Parts of the Eye and How We See - American Academy of Ophthalmology. *American Academy of Opthalmology Website*, 1–4.

Brifcani, A. M. A., & Al-Bamerny, J. N. (2010). Image compression analysis using multistage vector quantization based on discrete wavelet transform. *Proceedings of 2010 International Conference on Methods and Models in Computer Science, ICM2CS-2010*, 46–53. https://doi.org/10.1109/ICM2CS.2010.5706717

By, E. (2018). Deep Learning in Biometrics. In *Deep Learning in Biometrics*. https://doi.org/10.1201/b22524

Cantoni, V., Cascone, L., Nappi, M., & Porta, M. (2020). Demographic classification through pupil analysis. *Image and Vision Computing*, 102, 103980. https://doi.org/10.1016/J.IMAVIS.2020.103980

Cascone, L., Medaglia, C., Nappi, M., & Narducci, F. (2020). Pupil size as a soft biometrics for age and gender classification. *Pattern Recognition Letters*, 140, 238–244. https://doi.org/10.1016/j.patrec.2020.10.009

Czajka, A., & Becker, B. (2019). Application of dynamic features of the pupil for iris presentation attack detection. *Advances in Computer Vision and Pattern Recognition*, 151–168. https://doi.org/10.1007/978-3-319-92627-8_7

Daugman, J. (2009). How Iris Recognition Works. *The Essential Guide to Image Processing*, 14(1), 715–739. https://doi.org/10.1016/B978-0-12-374457-9.00025-1

Duan, M., Li, K., Yang, C., & Li, K. (2018). A hybrid deep learning CNN–ELM for age and gender classification. *Neurocomputing*, 275, 448–461. https://doi.org/10.1016/j.neucom.2017.08.062

Eskandari, M., & Sharifi, O. (2019). Effect of face and ocular multimodal biometric systems on gender classification. *IET Biometrics*, 8(4), 243–248. https://doi.org/10.1049/iet-bmt.2018.5134

Fang, B., Lu, Y., Zhou, Z., Li, Z., Yan, Y., Yang, L., Jiao, G., & Li, G. (2019). Classification of genetically identical left and right irises using a convolutional neural network. *Electronics (Switzerland)*, 8(10), 2–11. https://doi.org/10.3390/electronics8101109

Feng, G., & Wu, Y. (2008). An iris recognition algorithm based on DCT and GLCM. *Optical and Digital Image Processing*, 7000, 70001H. https://doi.org/10.1117/12.780158

Gu, S., & Ding, L. (2019). A Complex-Valued VGG Network Based Deep Learing Algorithm for Image Recognition. *9th International Conference on Intelligent Control and Information Processing, ICICIP 2018*, 340–343. https://doi.org/10.1109/ICICIP.2018.8606702

Hara, K., Saito, D., & Shouno, H. (2015). Analysis of function of rectified linear unit used in deep learning. *Proceedings of the International Joint Conference on Neural Networks*, 2015-Septe. https://doi.org/10.1109/IJCNN.2015.7280578

Hassan, M. M., Hussein, H. I., Eesa, A. S., & Mstafa, R. J. (2021). Face recognition based on gabor feature extraction followed by fastica and lda. *Computers, Materials and Continua*, 68(2), 1637–1659. https://doi.org/10.32604/CMC.2021.016467

Hinton, G. E., Srivastava, N., Krizhevsky, A., Sutskever, I., & Salakhutdinov, R. R. (2012). *Improving neural networks by preventing co-adaptation of feature detectors*.

Hollingsworth, K., Bowyer, K. W., & Flynn, P. J. (2009). Pupil dilation degrades iris biometric performance. *Computer Vision and Image Understanding*, 113(1), 150–157. https://doi.org/10.1016/J.CVIU.2008.08.001

JG Daugman - US Patent 5, 291,560, & 1994, undefined. (n.d.). Biometric personal identification system based on iris analysis. *Google Patents*.

Juefei-Xu, F., & Savvides, M. (2012). Unconstrained periocular biometric acquisition and recognition using COTS PTZ camera for uncooperative and non-cooperative subjects. *Proceedings of IEEE Workshop on Applications of Computer Vision*, 201–208. https://doi.org/10.1109/WACV.2012.6163051

Kazakov, T. (2011). *Iris Detection and Normalization*.

Khalifa, I. A., Zeebaree, S. R. M., Ataş, M., & Khalifa, F. M. (2019). Image Steganalysis in Frequency Domain Using Co-Occurrence Matrix and Bpnn. *Science Journal of University of Zakho*, 7(1), 27–32. https://doi.org/10.25271/SJUOZ.2019.7.1.574

Khalifa, N. E. M., Taha, M. H. N., Hassanien, A. E., & Mohamed, H. N. E. T. (2019). Deep iris: Deep learning for gender classification through iris patterns. *Acta Informatica Medica*, 27(2), 96–102. https://doi.org/10.5455/aim.2019.27.96-102

Khan, A. R., Doosti, F., Karimi, M., Harouni, M., Tariq, U., Fati, S. M., & Ali Bahaj, S. (2021). Authentication through gender classification from iris images using support vector machine. *Microscopy Research and Technique*, 84(11), 2666–2676. https://doi.org/10.1002/jemt.23816

Khan, M. T., Arora, D., & Shukla, S. (2013). Feature extraction through iris images using 1-D Gabor filter on different iris datasets. *2013 6th International Conference on Contemporary Computing, IC3 2013*, 445–450. https://doi.org/10.1109/IC3.2013.6612236

Kim, D. H. (2020). *Deep Convolutional GANs for Car Image Generation*.

Koklu, M., & Ozkan, I. A. (2020). Multiclass classification of dry beans using computer vision and machine learning techniques. *Computers and Electronics in Agriculture*, 174(June 2019), 105507. https://doi.org/10.1016/j.compag.2020.105507

Krizhevsky, A., Sutskever, I., & Hinton, G. E. (2012). ImageNet classification with deep convolutional neural networks. *Advances in Neural Information Processing Systems*, *2*, 1097–1105.

Kumar, D. R. S., Raja, K. B., Nuthan, N., Sindhuja, B., Supriya, P., Chhotaray, R. K., & Pattnaik, S. (2011). Iris recognition based on DWT and PCA. *Proceedings - 2011 International Conference on Computational Intelligence and Communication Systems, CICN 2011*, 489–493. https://doi.org/10.1109/CICN.2011.102

Kumar, S., Singh, S. K., Abidi, A. I., Datta, D., & Sangaiah, A. K. (2017). Group Sparse Representation Approach for Recognition of Cattle on Muzzle Point Images. *International Journal of Parallel Programming 2017 46:5*, *46*(5), 812–837. https://doi.org/10.1007/S10766-017-0550-X

Kumari, S., Bakshi, S., & Majhi, B. (2012). Periocular Gender Classification using Global ICA Features for Poor Quality Images. *Procedia Engineering*, *38*, 945–951. https://doi.org/10.1016/J.PROENG.2012.06.119

Li, H., Yue, X., Wang, Z., Wang, W., Tomiyama, H., & Meng, L. (2021). A survey of Convolutional Neural Networks —From software to hardware and the applications in measurement. *Measurement: Sensors*, *18*, 100080. https://doi.org/10.1016/J.MEASEN.2021.100080

Luo, Z. (2012). Iris Feature Extraction and Recognition Based on Wavelet-Based Contourlet Transform. *Procedia Engineering*, *29*, 3578–3582. https://doi.org/10.1016/J.PROENG.2012.01.534

Mabuza-Hocquet, G., Ngejane, C. H., & Lefophane, S. (2018). Predicting and Classifying Gender from the Human Iris: A Survey on Recent Advances. *2018 International Conference on Advances in Big Data, Computing and Data Communication Systems, IcABCD 2018*, 1–5. https://doi.org/10.1109/ICABCD.2018.8465471

Manyala, A., Cholakkal, H., Anand, V., Kanhangad, V., & Rajan, D. (2019). CNN-based gender classification in near-infrared periocular images. *Pattern Analysis and Applications*, *22*(4), 1493–1504. https://doi.org/10.1007/s10044-018-0722-3

Patil, A., R, K., & Gornale, S. (2019). Analysis of Multi-modal Biometrics System for Gender Classification Using Face, Iris and Fingerprint Images. *International Journal of Image, Graphics and Signal Processing*, *11*(5), 34–43. https://doi.org/10.5815/ijigsp.2019.05.04

Payasi, M., & Cecil, K. (2021). LBP and Iris Features based Human Gender Classification using radial Support Vector Machine. *2021 4th International Conference on Electrical, Computer and Communication Technologies, ICECCT 2021*. https://doi.org/10.1109/ICECCT52121.2021.9616923

Podder, P., Rubaiyat Hossain Mondal, M., & Kamruzzaman, J. (2022). Iris feature extraction using three-level Haar wavelet transform and modified local binary pattern. *Applications of Computational Intelligence in Multi-Disciplinary Research*, 1–15. https://doi.org/10.1016/B978-0-12-823978-0.00005-8

Rahim, Z., Kadhim, H., & Salih, M. (2021). Survey of Iris Recognition using Deep Learning Techniques. *Journal of Al-Qadisiyah for Computer Science and Mathematics*, *13*(3), 47–56.

Rai, P., & Khanna, P. (2012). Gender classification techniques: A review. *Advances in Intelligent and Soft Computing*, *166 AISC*(VOL. 1), 51–59. https://doi.org/10.1007/978-3-642-30157-5_6

Rajput, M., & Sable, G. (2020). Deep Learning Based Gender and Age Estimation from Human Iris. *SSRN Electronic Journal*, 1–9. https://doi.org/10.2139/ssrn.3576471

Ramón-Balmaseda, E., Lorenzo-Navarro, J., & Castrillón-Santana, M. (2012). Gender classification in large databases. *Lecture Notes in Computer Science (Including Subseries Lecture Notes in Artificial Intelligence and Lecture Notes in Bioinformatics)*, *7441 LNCS*, 74–81. https://doi.org/10.1007/978-3-642-33275-3_9

Rattani, A., Reddy, N., & Derakhshani, R. (2018). Convolutional neural networks for gender prediction from smartphone-based ocular images. *IET Biometrics*, *7*(5), 423–430. https://doi.org/10.1049/iet-bmt.2017.0171

Rekha, V., Gurupriya, S., Gayadhri, S., & Sowmya, S. (2019). Dactyloscopy based gender classification using machine learning. *2019 IEEE International Conference on System, Computation, Automation and Networking, ICSCAN 2019*, 1–5. https://doi.org/10.1109/ICSCAN.2019.8878756

Reshma, P. A., Divya, K. V., Therattil, G. J., & Subair, T. B. (2018). A study of gender recognition from Iris: A literature survey. *Proceedings of the International Conference on Intelligent Sustainable Systems, ICISS 2017*, *December 2017*, 888–891. https://doi.org/10.1109/ISS1.2017.8389306

Rinky, B. P., Mondal, P., Manikantan, K., & Ramachandran, S. (2012). DWT based Feature Extraction using Edge Tracked Scale Normalization for Enhanced Face Recognition. *Procedia Technology*, *6*, 344–353. https://doi.org/10.1016/j.protcy.2012.10.041

Sharanappa Gornale, S., Patil, A., & Ramchandra, K. (2020). Multimodal Biometrics Data Analysis for Gender Estimation Using Deep Learning. *International Journal of Data Science and Analysis*, *6*(2), 64. https://doi.org/10.11648/j.ijdsa.20200602.11

Sharma, N., Jain, V., & Mishra, A. (2018). An Analysis Of Convolutional Neural Networks For Image Classification. *Procedia Computer Science*, *132*, 377–384. https://doi.org/10.1016/J.PROCS.2018.05.198

Szegedy, C., Liu, W., Jia, Y., Sermanet, P., Reed, S., Anguelov, D., Erhan, D., Vanhoucke, V., & Rabinovich, A. (2014). Going Deeper with Convolutions. *Proceedings of the IEEE Computer Society Conference on Computer Vision and Pattern Recognition*, *07-12-June*, 1–9. https://doi.org/10.48550/arxiv.1409.4842

Tapia, J., & Arellano, C. (2019). Gender Classification from Iris Texture Images Using a New Set of Binary Statistical Image Features. *2019 International Conference on Biometrics, ICB 2019*. https://doi.org/10.1109/ICB45273.2019.8987245

Tapia, J., Arellano, C., & Viedma, I. (2019). Sex-classification from cellphones periocular iris images. *Advances in Computer Vision and Pattern Recognition*, 227–242. https://doi.org/10.1007/978-3-030-26972-2_11

Tapia, J. E., & Perez, C. A. (2019). Gender Classification from NIR Images by Using Quadrature Encoding Filters of the Most Relevant Features. *IEEE Access*, *7*, 29114–29127. https://doi.org/10.1109/ACCESS.2019.2902470

Vijayalaxmi, & Rao, P. S. (2012). Eye detection using Gabor Filter and SVM. *International Conference on Intelligent Systems Design and Applications, ISDA*, 880–883. https://doi.org/10.1109/ISDA.2012.6416654

Wildes, R. P. (1997). Iris recognition: An emerging biometric technology. *Proceedings of the IEEE*, *85*(9), 1348–1363. https://doi.org/10.1109/5.628669

Willoughby, C. E., Ponzin, D., Ferrari, S., Lobo, A., Landau, K., & Omidi, Y. (2010). Anatomy and physiology of the human eye: effects of mucopolysaccharidoses disease on structure and function – a review. *Clinical & Experimental Ophthalmology*, *38*(SUPPL. 1), 2–11. https://doi.org/10.1111/J.1442-9071.2010.02363.X

Wu, C., Wen, W., Afzal, T., Zhang, Y., Chen, Y., & Li, H. H. (2017). A compact DNN: Approaching GoogLeNet-level accuracy of classification and domain adaptation. *Proceedings - 30th IEEE Conference on Computer Vision and Pattern Recognition, CVPR 2017*, *2017-January*, 761–770. https://doi.org/10.1109/CVPR.2017.88